\documentclass[nohyperref]{article}

\usepackage{microtype}
\usepackage{graphicx}
\usepackage{subfigure}
\usepackage{booktabs} 

\usepackage{layouts}
\usepackage[table,xcdraw]{xcolor}

\usepackage{hyperref}

\usepackage[accepted]{icml2022}

\usepackage{amsmath}
\usepackage{amssymb}
\usepackage{mathtools}
\usepackage{amsthm}
\usepackage{ mathrsfs }
\usepackage[inline]{enumitem}

\usepackage[capitalize,noabbrev]{cleveref}

\usepackage{pgf}

\theoremstyle{plain}
\newtheorem{theorem}{Theorem}[section]

\theoremstyle{definition}

\theoremstyle{remark}
\newtheorem{remark}[theorem]{Remark}

\usepackage[textsize=tiny]{todonotes}

\DeclareMathOperator*{\argmin}{arg\,min}
\DeclareMathOperator{\st}{s.t.}

\DeclareMathOperator{\EX}{\mathbb{E}}
\newcolumntype{M}[1]{>{\centering\arraybackslash}m{#1}}

\graphicspath{{./figures/}}

\icmltitlerunning{MPC-IL for Safe and Human-like AVs}

\begin{document}

\twocolumn[
\icmltitle{MPC-based Imitation Learning for Safe and Human-like Autonomous Driving}



\icmlsetsymbol{equal}{*}

\begin{icmlauthorlist}
\icmlauthor{Flavia Sofia Acerbo}{siemens,meco,esat}
\icmlauthor{Jan Swevers}{meco,fm}
\icmlauthor{Tinne Tuytelaars}{esat}
\icmlauthor{Tong Duy Son}{siemens}
\end{icmlauthorlist}

\icmlaffiliation{siemens}{Siemens Digital Industries Software, Leuven, Belgium}
\icmlaffiliation{meco}{KU Leuven, Mechanical Engineering Department, Leuven, Belgium}
\icmlaffiliation{esat}{KU Leuven, Electrical Engineering Department, Leuven, Belgium}
\icmlaffiliation{fm}{DMMS-M core-lab Flanders Make, Leuven, Belgium}
\icmlcorrespondingauthor{Flavia Sofia Acerbo}{flavia.acerbo@siemens.com}

\icmlkeywords{Safe Imitation Learning, Model Predictive Control}

\vskip 0.3in
]



\printAffiliationsAndNotice{}  

\begin{abstract}
To ensure user acceptance of autonomous vehicles (AVs), control systems are being developed to mimic human drivers from demonstrations of desired driving behaviors. Imitation learning (IL) algorithms serve this purpose, but struggle to provide safety guarantees on the resulting closed-loop system trajectories.
On the other hand, Model Predictive Control (MPC) can handle nonlinear systems with safety constraints, but realizing human-like driving with it requires extensive domain knowledge. 
This work suggests the use of a seamless combination of the two techniques to learn safe AV controllers from demonstrations of desired driving behaviors, by using MPC as a differentiable control layer within a hierarchical IL policy. With this strategy, IL is performed in closed-loop and end-to-end, through parameters in the MPC cost, model or constraints. Experimental results of this methodology are analyzed for the design of a lane keeping control system, learned via behavioral cloning from observations (BCO), given human demonstrations on a fixed-base driving simulator.
\end{abstract}

\section{Introduction}
Currently, the design of autonomous vehicles (AVs) is primarly focused on safety. However, to ensure a broad use by the general public, AVs should also be comfortable and enjoyable. Comfort in AVs is a complex and multifaceted entity, concerning attributes such as motion sickness, perceived safety, understandability, trust and many others \cite{Elbanhawi2015}. Hence, comfort optimization covers different areas of AVs research, such as personalized driving controllers, novel cabin and seat design, active integrated chassis control or motion sickness minimization. In this work, we consider \emph{human-like driving}, a solution to improve AVs comfort which entails an implicit personalization of AVs behavior inspired by human driving behaviors \cite{Bellem2016, Kalabic2019}.
This can be realized in a model-based or in a data-driven approach; intensive research has been done on human driving models, with sensory, cognitive and muscular level of detail \cite{Nash2019}.  
Though accurate, these models are usually very task-specific and their identification for different tasks is expensive \cite{Kolekar2020}. Therefore, only a relatively small part of research on human-like AVs is implementing human driving models in the controllers, 
e.g. \cite{Lazcano2021}. For higher level of autonomy, research is primarly focusing on data-driven approaches, such as imitation learning (IL), which can obtain policies mimicking driving demonstrations experienced or labeled by the user as adequate.
In this domain, various solutions from both academia and industry \cite{Hawke2020} show great potential on learning policies end-to-end, i.e. from sensory data to control actions or states.

Imitation learning solutions for human-like driving usually involve deep learning (DL) in their architecture, and this limits the verifiability of the safety and stability properties of the closed-loop system. To provide guarantees, many works have proposed a hierarchical framework, where the DL components are combined with safety filters, frequently model-based controllers. In \cite{Pan2017, Chen2019, Pulver2021, Acerbo2020} such DL components represent trajectories fed to tracking controllers, but learning is done without directly considering their effect, hence disrupting the end-to-end learning. 
In the majority of these works, model predictive control (MPC) is used, due to its ability to explicitly handle hard constraints. Moreover, MPC is a generic controller, which can be designed to have parameters in its cost function, model or constraints. In this regard, there have been recent advances on the use of Model Predictive Control (MPC), as a differentiable policy in the context of IL \cite{Amos2018} and reinforcement learning (RL) \cite{Gros2020}. 

This work is a preliminary study to extend existing hierarchical structures of IL for AVs to differentiable MPC, which allows for end-to-end learning while guaranteeing safety and stability of the controlled system. Specifically, MPC is used as a reactive control layer and its parameters are provided by other DL indirect controllers, which map from sensory data to the parameters. 
The paper is structured as follows: first, we provide an introduction on related works concerning IL for AVs and previous studies on MPC with IL and RL. Then, we present the generic formulation of the proposed approach, namely MPC-IL, which leverages the MPC differentiability and the hierarchical decomposition. After that, we detail the approach under the behavioral cloning from observations algorithm (MPC-BCO) and finally we provide preliminary experimental results of MPC-BCO applied to the design of a human-like lane keeping controller from human demonstrations.

\section{Background and Related Work}
\subsection{Imitation Learning for Autonomous Driving} \label{backgroundIL}
Let us consider a fully observable Markov Decision Process characterized by a set of continuous states $s \in \mathcal{S} \subset \mathbb{R}^{n_s}$, which can be controlled with a set of continuous actions $a \in \mathcal{A} \subset \mathbb{R}^{n_a}$ and with a state transition probability distribution $s_{t+1} \sim P(\cdot| s_t, a_t)$. 
An agent can control the system with a policy $\pi(a_t|s_t)$. We denote the \emph{state occupancy measure} $\rho_\pi(s)$ induced by a policy $\pi$ as the density of occurrence of states, while following that policy, over an infinite time horizon, discounted by the factor $\gamma$: 
\(
	\rho_{\pi}^s = \sum_{t=0}^{\infty} \gamma^t P(s | \pi).
\)
Given an expert policy $\pi_E$ to be imitated, IL tries to find a policy $\pi$ minimizing the occupancy measure distance formulated as:
\begin{equation}
	\min_{\pi} \EX_{s \sim \rho_{\pi}^s }[\mathcal{L}(\pi_E,\pi)] = \mathcal{L}(\rho_{\pi_E}^s, \rho_{\pi}^s)\;.
\end{equation}
As demonstrated by \cite{Ghasemipour2019}, all IL algorithms can be read under this formulation. However, in the context of AVs, behavioral cloning (BC) is the algorithm with the most promising history of success \cite{Bansal2018, Scheel2021}.
Standard BC provides a straightforward solution to the IL problem, by learning the policy in a supervised way through direct maximum likelihood, i.e. by minimizing the distance between the action distributions under the expert state occupancy measure, as: \(\min_{\pi} \EX_{s \sim \rho_{\pi_E}^s }[\mathcal{L}(\pi_E,\pi)]\). This suffers from the covariate shift problem, since the policy $\pi$ generates a different occupancy measure when run in closed-loop, and then becomes unpredictable in its own induced distribution of states. Among different solutions to mitigate this problem, it has been shown by \cite{Ghasemipour2019}, that learning the marginal state distribution of the expert in addition to the action distribution, can heavily influence the results. Hence, behavioral cloning from observations (BCO) \cite{Torabi2018} aims at minimizing the Kullback–Leibler (KL) divergence on the state occupancy measures as:
\(\min_{\pi} \text{KL}(\rho_{\pi_E}^s||\rho_{\pi}^s).\) 
Nevertheless, BCO is more expensive than BC, since it requires to run simulations in the learning loop and needs solutions to estimate the policy gradient from the states.

\subsection{Model Predictive Control and Reinforcement Learning}
Model predictive control (MPC) is becoming a popular technique in AV research and industry. It is an optimal controller with receding horizon denoted in its most general form as:
\begin{equation}
	\begin{aligned}[t]
		a = \argmin_{x,u}  \quad & \sum_{k=0}^{N-1} l(\theta, x_k, u_k) + l_N(\theta, x_k, u_k) \\
		\st \quad & x_0 = \hat{x}\\
		& x_{k+1} = f(x_k, u_k, \theta)\\
		& h(x_k, u_k, \theta) \leq 0 \\
		& a = u_0 \;.
	\end{aligned}
\end{equation}
At each time step, the current state of the system is estimated as $\hat{x}$. Then, MPC minimizes a cost function across a prediction horizon $N$, where the evolution of the model is simulated with $f$ and where its states and controls should not violate the constraints specified by $h$. Usually, when talking about learning-based MPC, we refer to system identification of the underlying model $f$, given data from the controlled system. However, it is possible to parametrize also its cost function and constraints.
In this regard, MPC was recently proposed as a solution for safety and sample efficiency issues in IL and RL. \cite{Gros2020} showed that using a highly parametric MPC as a differentiable function approximator for RL results in safe and stable RL policies and allows MPC to optimize its closed-loop performance, even when its underlying model $f$ is wrong and system identification is not performed. \cite{Amos2018} have benchmarked the use of differentiable MPC for classical IL/RL problems showing the advantages in terms of sample efficiency and learning flexibility with respect to generic IL and system identification. However, these studies did not consider learning MPC parameters in the form of goals, references or other generic complex behavior to be imitated and therefore considered MPC as the entire policy, without combining it with other learning components.

\section{MPC-based Imitation Learning}
This section presents the formulation and generic algorithmic framework of our MPC-IL method. Here, MPC is used as the differentiable final control layer of a hierarchical policy. This means that MPC parameters are given by preceding indirect controllers, which are learnt with end-to-end imitation learning. Moreover, a specific IL algorithm, namely behavioral cloning from observations (BCO), is detailed as an example of the MPC-IL approach. 

\subsection{Problem Formulation}
As defined in Section \ref{backgroundIL} we consider IL as the minimization of the occupancy measure distance, according to a certain metric $\mathcal{L}$. The system under study is controlled by the human with an (unknown) stochastic policy $\pi_H$ as $a_t \sim \pi_H(\cdot|s_t)$, while the mimicking controller controls it with a deterministic policy parametrized by $\theta$ as $a_t = \pi_\theta(s_t)$. 
We now formulate this problem in the context where the controller part of the policy is represented by a MPC, optimizing its control action $a$ according to the state $s$ of the system and the parameters $\theta$ that are part of the objective function, model and constraints:
\begin{equation}\label{eq:mpcil}
	\begin{aligned}
		\min_{\theta} \quad & \EX_{s \sim \rho_{\pi_\theta}^s }[\mathcal{L}(\pi_H,\pi_\theta)] = \mathcal{L}(\rho_{\pi_H}^s, \rho_{\pi_\theta}^s)\\
		\text{with} \quad & \pi_\theta \in 
		\begin{aligned}[t]
			\argmin_{x,u}  \quad & \sum_{k=0}^{N-1} l(\theta, x_k, u_k) + l_N(\theta, x_k, u_k) \\
			\st \quad  & x_0 = s\\
			& x_{k+1} = f(x_k, u_k, \theta)\\
			& h(x_k, u_k, \theta) \leq 0 \\
			& a = u_0 \;.
		\end{aligned}
	\end{aligned}
\end{equation}
Equation (\ref{eq:mpcil}) describes a \emph{bilevel optimization problem}, where the upper-level problem is the imitation of policy $\pi$ with respect to the human policy $\pi_H$ and the lower-level problem in the MPC optimization, given a state $s$. 
Among gradient-based solution techniques it is common to replace the lower level optimization problems with their Karush-Kuhn-Tucker (KKT) conditions as constraints \cite{Hatz2012}. 
However, directly solving a constrained optimization problem is not trivial in the context of imitation and reinforcement learning. Moreover, it requires second order information (i.e. Hessian matrices) and that does not scale well when neural networks and large amount of data are involved.
Conversely, the adopted solution is based on first-order information only \cite{Amos2018}. The core idea is to compute the gradient of the parametrized solution of the MPC with respect to the optimization variables of the upper-level imitiation problem, and then solve it as an unconstrained one with gradient descent steps of size $\alpha$ as:
\begin{equation} \label{eq:gd}
	\begin{aligned}
		\theta & \leftarrow \theta -\alpha\left[\frac{\partial{\mathcal{L}}}{\partial{\theta}}\right]^T \\ \left[\frac{\partial{\mathcal{L}}}{\partial{\theta}}\right]^T &=
		\left[\frac{\partial{\pi_\theta}}{\partial{\theta}}\right]^T \left[\frac{\partial{\mathcal{L}}}{\partial{\rho_{\pi_\theta}^s}}
		\frac{\partial{\rho_{\pi_\theta}^s}}{\partial{\pi_\theta}}\right]^T.
	\end{aligned}
\end{equation}
The computation of $\left[\frac{\partial{\mathcal{L}}}{\partial{\theta}}\right]^T$ requires:
\begin{itemize}
	\item $\left[\frac{\partial{\mathcal{L}}}{\partial{\rho_{\pi_\theta}^s}}
	\frac{\partial{\rho_{\pi_\theta}^s}}{\partial{\pi_\theta}}\right]^T$, which depends on the adopted IL algorithm, defining the distance $\mathcal{L}$ and how to approximate the occupancy measure $\rho$.
	\item $\left[\frac{\partial{\pi_\theta}}{\partial{\theta}}\right]^T$, i.e. the derivative of the MPC problem at its (locally) optimal solution. In the next section we describe how to efficiently compute it via automatic differentiation.
\end{itemize}

\subsection{Backpropagation through the MPC Problem}
We now detail how to compute the gradient of a policy $\pi_\theta$ with a MPC control layer. Let us introduce the Lagrangian associated to the MPC problem of Equation (\ref{eq:mpcil}):
\begin{equation}
	\begin{aligned}
		\mathscr{L}(x, u,\mu, \lambda, \theta) = \quad &\sum_{k=0}^{N-1} l(\theta, x_k, u_k) + l_N(\theta, x_k, u_k) \\ 
		+ & \mu^Th(x,u,\theta) + \lambda_0^T (x_0 - s) \\
		+ &\sum_{k=0}^{N-1} \lambda^T_{k+1}\left[ f(x_k, u_k, \theta) - x_{k+1} \right],
	\end{aligned}
\end{equation}
where $\mu, \lambda$ are the multipliers associated to the dual solution. For simplicity of notation, we denote as $z = \left[ x, u, \lambda, \mu \right]$ the vector of the primal and dual variables associated with the optimal solution, each of them with $N$ components e.g., $x_0, x_1, \dots, x_N$. At a local optimum $z^*$ and if the problem satisfies the linear independence constraint qualification (LICQ) and the second-order sufficient conditions (SOCS) \cite{Gros2020}, then the following Karush–Kuhn–Tucker (KKT) conditions hold:
\begin{align}
	& \nabla_{x,u} \mathscr{L}(z^*, \theta) = 0  \label{stationarity}  \\
	& \begin{aligned} \label{primalfeas}
		&x^*_0 = s,  \\
		&x^*_{k+1} = f(x^*_k, u^*_k, \theta) \; \text{for} \: k=0,...,N-1, \\
		&h(z^*, \theta) \leq 0
	\end{aligned}\\
	& \mu^* \geq 0 \label{dualfeas} \\
	& \mu^{*T}h(z^*,\theta) = 0. \label{complslack}
\end{align}
If the active set of the inequalities constraints at $z^*$ is known, then we can write the previous equations as an implicit function $F(z^*,\theta) \leftrightarrow z^* = \pi_\theta$ as:
\begin{equation}
	F(z^*,\theta) = \begin{bmatrix}
		\nabla_{x,u} \mathscr{L}(z^*, \theta)\\
		I_hh(z^*,\theta) - I_h^0\mu^*\\
		\mathbf{f}(z^*, \theta)
	\end{bmatrix}
= 0\;,
\end{equation}
where $I_h$ and $I_h^0$ are diagonal matrices denoting active and inactive inequality constraints, respectively. $\mathbf{f}$ represents the matrix form of the $N$ equality constraints from Equation (\ref{primalfeas}).
According to the implicit function theorem, we can write the Jacobian of $\pi_\theta$ as:
\begin{equation}
	\frac{\partial \pi_\theta}{\partial \theta} = -\frac{\partial F}{\partial z}^{-1}\frac{\partial F}{\partial \theta}\;.
\end{equation}
The full $\frac{\partial \pi_\theta}{\partial \theta}$ may be very expensive to compute. However, we are interested in the Jacobian-times-vector product
\( \left[\frac{\partial{\pi_\theta}}{\partial{\theta}}\right]^T \bar{z}\). 
This is also known as adjoint sensitivity and can be computed efficiently within automatic differentiation (AD) engines at the cost of a linear system solve and a reverse mode sweep \cite{Andersson2018} as:
\(
	\frac{\partial \pi_\theta}{\partial \theta} \bar{z} = -\frac{\partial F}{\partial \theta}^{T} \frac{\partial F}{\partial z}^{-T} \bar{z}\;.
\)

\remark
In order to compute the gradient we must ensure that $\frac{\partial F}{\partial z}$ is invertible. This means that, alongside the conditions of LICQ and SOCS, it is required that the optimal solution $z^*$ where we compute the gradient is at least locally unique. This is not guaranteed in the case of nonlinear MPC and therefore we should impose some sort of regularization on the primal variables.

\subsection{Hierarchical Decomposition}\label{hierarchical}
\begin{figure}[!tb]
	\centering
	\def\svgwidth{\columnwidth}
	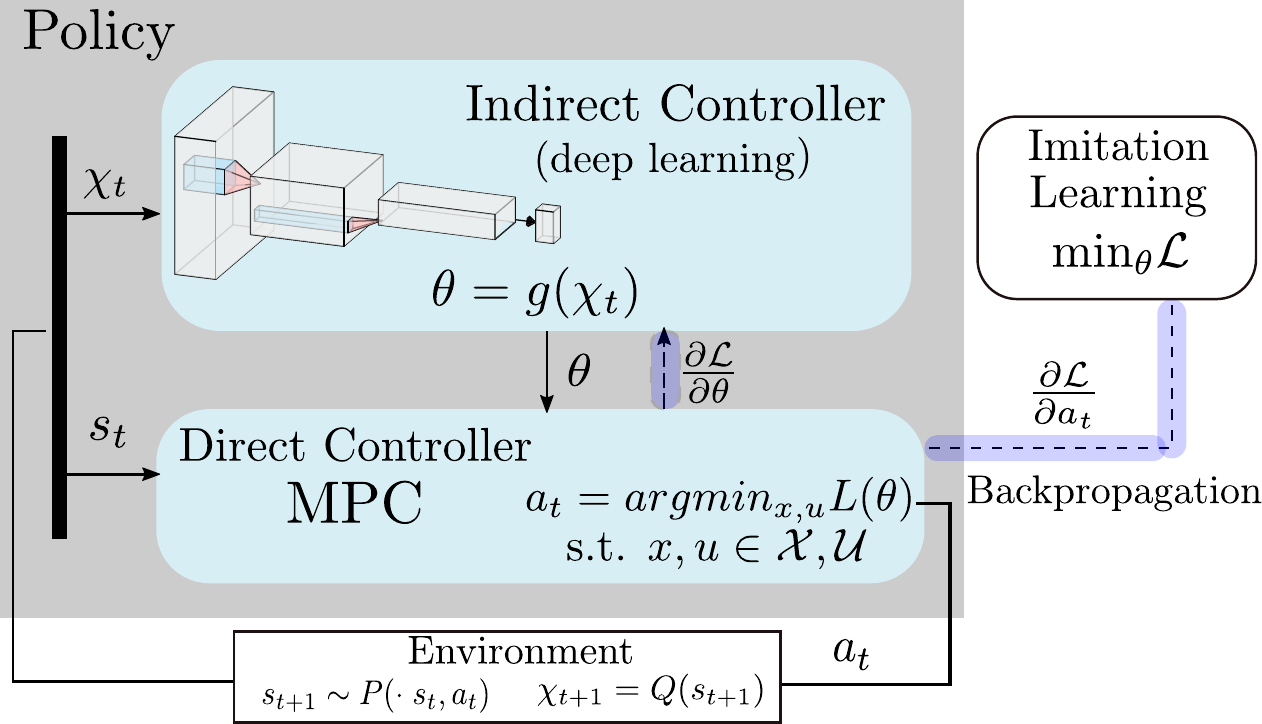
	\caption{Hierarchical decomposition of the policy with a MPC control layer: the MPC provides the mapping $s_t \rightarrow a_t$, based on a constrained optimization. The indirect controller outputs the MPC parameters $\theta$ based on the latent variable $\chi_t$. The dashed line show the flow of the gradient through the whole policy, possible thanks to the MPC differentiability.}
	\label{fig:hierarchicaldiagram}
\end{figure}

In the available literature the parameters $\theta$ refer to physical properties in the model or to the weight matrices in a quadratic regulator cost, and are therefore static \cite{Amos2018}. Nevertheless, to learn more complex and strategic behaviors, as it is the case for end-to-end AV control from demonstrations, we suggest the need to combine MPC control with other differentiable layers, as graphically represented in Figure \ref{fig:hierarchicaldiagram}. Let us consider some context information about the system at time $t$ in the form of a latent variable $\chi_t$, inferrable from the system current state as $\chi_t = Q(s_t)$; for example, $\chi_t$ may be a high-dimensional input (e.g. camera images), from which it would be convenient to learn end-to-end. For this purpose, we suggest to learn a function $\theta = g (\chi_t)$ that maps the latent variable $\chi_t$ to the MPC parameters $\theta$. For example, $\theta$ can represent the goal state of the system or a trajectory to follow, which changes depending on the current situation at time $t$; We denote this approach as \emph{hierarchical decomposition}; the MPC is responsible for the reactive control behavior, i.e. the mapping $s_t \rightarrow a_t$, while the preceding layers learn subtasks that indirectly influence the closed-loop behavior of the system. Such layers are said to be \emph{indirect controllers}. Then, the policy is written as: \(a_t = \pi_{\theta(\chi_t)}(s_t)\).


\subsection{MPC-based Behavioral Cloning from Observations (MPC-BCO)} \label{mpcbco}
It is now detailed how we use the MPC-IL method to learn a deterministic hierarchical policy $\pi_\theta$ with a MPC control layer using closed-loop policy learning via behavioral cloning from observations (BCO) and a differentiable simulator.

\paragraph{Algorithm}
We consider rollout state trajectories of length $T$, induced by the policy $\pi_\theta$: $\tau_{\pi_\theta} = s_0, s_1, \dots, s_{T}$, and state trajectories sampled from the human demonstrations: $\tau_{\pi_H} = s_0^*, s_1^*, \dots, s_{T}^*$. We frame the problem as the minimization of the L2 state error $L(s_t, s_t^*) = ||s_t - s_t^*||_2^2$ between the human and policy on sequences of induced states, when starting from the same initial state $s_0^*$. These sequences are acquired through a simulation model $F$, providing the state transitions $s_{t+1} = F(s_t, a_t)$. 
Let us assume that the first derivative of $F$ around a certain  $s_t, a_t$ can be computed (this assumption is clarified in the last paragraph of this section). Therefore, we suggest computing $\nabla_\theta L(s_t)$ with backpropagation through time (BPTT) as:
\begin{equation}
\begin{aligned}
			\nabla_\theta L_{| s = s_t} &= \nabla_s L_{| s = s_t} \left(\right.
		\nabla_a F \nabla_\theta a_{| a = a_{t-1}} + \\
		& + \nabla_a F \nabla_s a \nabla_\theta s_{| a = a_{t-1}, s = s_{t-1}} + \\
		&+ \nabla_s F \nabla_\theta s_{| s = s_{t-1}} \left.\right)\;.
\end{aligned}
\end{equation}
Hence, with BPTT future state errors are accounted by the earlier actions taken by the policy.

We define $J(\theta) = \EX_{s_t \sim \rho_{\pi_\theta}^s } \EX_{s_t^* \sim \rho_{\pi_H}^s } \left[ \sum_{t=0}^{T} L(s_t, s_t^*) \right]$ the sum of the L2 pose errors along a trajectory.
Then, to update the policy, we compute the derivatives of $J$ over $(s_t, a_t, s_{t+1})$ transitions with BPTT, leveraging on the differentiability of the policy and of the simulator. From \cite{Heess2015}, the gradient is computed by the following recursive formulas:
\begin{equation}\label{eq:recursivegrad}
	\begin{aligned}
		\frac{\partial J_t}{\partial s_t} &= \frac{\partial L}{\partial s_t} + \frac{\partial J_{t+1}}{\partial s_{t+1}} \left( \frac{\partial F}{\partial s_{t}} + \frac{\partial F}{\partial a_{t}} \frac{\partial \pi_{\theta}}{\partial s_{t}}_{|\chi_t}\right) \\ 
		\frac{\partial J_t}{\partial \theta} &= \frac{\partial J_{t+1}}{\partial s_{t+1}}\frac{\partial F}{\partial a_{t}}\frac{\partial \pi_{\theta}}{\partial \theta}_{|\chi_t} + \frac{\partial J_{t+1}}{\partial \theta}\;.
	\end{aligned}
\end{equation}
The final gradient $J(\theta)$ is computed by applying the equations from (\ref{eq:recursivegrad}) recursively, starting from $t = T$ up to $t = t_s$. We suggest to not backpropagate until $t_s = 0$ to avoid bias; since the initial state $s_0^*$ belongs to the human state occupancy metric, there may be a transitory part before the policy reaches a steady state corresponding to its actual occupancy metric. The complete algorithm is described in Algorithm \ref{alg:mpcbco}.

\remark
Note that in Equation (\ref{eq:recursivegrad}) the loss and its gradients are computed only with respect to the state $s$ and \emph{not with respect to the latent variable $\chi$}, which we provide to the indirect controllers of the policy. This is because we assume that $\chi$ is a nondifferentiable variable, coming from a black-box part of the simulation. As an example, we can consider $\chi$ as the result of an object detection. Since the policy actions cannot influence the object detection itself, it is useless to compute $\frac{\partial J_t}{\partial \chi_t}$. On the other hand, we may want to keep a specific distance to the specific detected object, which would be influenced by the policy actions and that would be included in the state $s_t$. 

\begin{algorithm}
	\caption{MPC-BCO}\label{alg:mpcbco}
	\begin{algorithmic}
		\STATE {\bfseries Input:} Human sampled trajectories $D(\tau_{\pi_H}) = \tau^*_1, \tau^*_2, \dots, \tau^*_m \sim \rho_{\pi_H}$ where $\tau^* = {s^*_0, s^*_1, \dots, s^*_T}$
		\STATE Initialize $\pi_{\theta}$ with $\theta_0$
		\FOR{$\tau^* \sim D(\tau_{\pi_H}) $}
		\STATE $s_0 = s^*_0$, $\chi_0 = Q(s^*_0)$
		\FOR{$t = 0$ {\bfseries to} $T-1$}
		\STATE \COMMENT{Policy Simulation Rollout}
		\STATE $a_t = \pi_{\theta(\chi_t)}(s_t)$
		\STATE $s_{t+1} = F(s_t, a_t)$
		\STATE $\chi_{t+1} = Q(s_{t+1})$
		\ENDFOR
		\STATE $\frac{\partial J_{T+1}}{\partial s_{T+1}} = \frac{\partial J_{T+1}}{\partial \theta} = 0$
		\FOR{$t = T$ {\bfseries down to} $t_s$}
		\STATE \COMMENT{BPTT}
		\STATE $\frac{\partial J_t}{\partial s_t} = \frac{\partial L}{\partial s_t} + \frac{\partial J_{t+1}}{\partial s_{t+1}} \left( \frac{\partial F}{\partial s_{t}} + \frac{\partial F}{\partial a_{t}}\frac{\partial \pi_{\theta}}{\partial s_{t}}\right)$
		\STATE $\frac{\partial J_t}{\partial \theta} = \frac{\partial J_{t+1}}{\partial s_{t+1}}\frac{\partial F}{\partial a_{t}}\frac{\partial \pi_{\theta}}{\partial \theta} + \frac{\partial J_{t+1}}{\partial \theta}$
		\ENDFOR
		\STATE $\theta \leftarrow \theta -\alpha\left[\frac{\partial J_{t_s}}{\partial{\theta}}\right]^T$ \COMMENT{Gradient Update}
		\ENDFOR
	\end{algorithmic}
\end{algorithm}


\paragraph{Backpropagation through the simulation model}
The MPC-BCO algorithm hinges on the first-order differentiability of the simulation model $s_{t+1} = F(s_t, a_t)$. Here, we describe two possibilities that enable this: differentiating through the simulator itself and differentiating through the MPC model.
Recently there have been interesting developments towards the implementation of differentiable simulators for autonomous driving, i.e. extensions of existing numerical solvers with the functionality to compute gradients with respect to their inputs. These simulators should consist of both differentiable traffic scenario and ego vehicle dynamics. As to the former, examples can be found in \cite{Suo2021} and \cite{Scheel2021}. As to the latter, high-fidelity vehicle dynamics models can be used without their explicit knowledge, as the only thing that's required is their evaluation for a given set of states and their respective Jacobians, which both are provided for example by the Functional Mock-up Interface (FMI) standard version 2.0 descriptions \cite{Blockwitz2012}.

If a differentiable simulator is not available, it is possible to exploit again the MPC structure. Indeed, given that the MPC relies on a good enough model, it is possible to simulate forward and backward by mapping $s,a \rightarrow x,u \Rightarrow x_{k+1} = f(x_k, u_k, \theta) \rightarrow s_{t+1} = f(s_t, u_t, \theta)$ and therefore compute a good approximation of the system gradients through $f$. However, as it can be seen from the previous formulation, the model can depend on the parameters $\theta$ which are not guaranteed to converge to the actual system parameter during learning, since we are optimizing only a closed-loop metric and not performing system identification. The possibility to combine system identification with imitation learning, similar to what is presented by \cite{Martinsen2020}, is beyond the scope of this paper and will be object of future research.

\section{MPC-BCO for Lane Keeping Control}
This section presents experimental results of the algorithm described in Section \ref{mpcbco}. The considered application is the design of a lane keeping control system, whose characteristics are learned from human demonstrations, collected on a fixed-base driving simulator. The MPC is designed as a path follower in Frenet coordinates and is considered as the control layer of a policy learned with Behavioral Cloning from Observations (BCO), in closed-loop. The following results are presented: learning of static parameters in the MPC stage cost and learning of a indirect controller for the MPC terminal cost. The latter shows capability in learning complex behavior and in adapting its output to the MPC dynamics to obtain the desired closed-loop trajectories.

\subsection{Human Curve Driving}
Although lane keeping may seem like a trivial automated control function to realize, it is quite hard to understand and realize it in a human-like way. Extensive research is available on human steering behaviors in curves. \cite{Godthelp1984} showed that during curve driving a measure related to a visual feature, namely the time to lane crossing (TLC), is always kept above a minimum threshold, characterizing different driving styles. The TLC is defined as: \(\text{TLC} = \frac{D}{V}\), where $V$ is the longitudinal speed of the vehicle and the distance to lane crossing $D$ is defined as the distance where the vehicle would exit the lane if no actions were taken from the current state.
To influence the TLC the driver can modify its speed $V$ and/or its distance to lane crossing $D$, which is determined by the lateral centerline deviation $d$ and heading error $\theta$, defined in Figure \ref{vehicle}.
As shown by \cite{Barendswaard2019}, the desired values of these variables change according to different phases of curve driving (estimation, anticipation, entry and exit) and the transitory behaviour before curve entry is often overlooked by steering controllers.
These considerations motivate the use of the MPC-BCO algorithm to learn complex behaviors in lane keeping controllers from human demonstrations.

\subsection{Human Demonstrations}\label{demos}
Human demonstrations of lane keeping control are collected using a fixed-base driving simulator. Firstly, the track and the visual feedback presented to the driver are designed with Simcenter Prescan. The track is 1200m long and made of 7 curved roads with different lengths and curvatures. 
Secondly, the ego vehicle is modelled via Simcenter Amesim as a 15DOF high-fidelity model of a Ford Focus. 
Finally, the two softwares are directly interfaced and can be run simultenously through a C++, Python or Simulink interface, to create the whole simulation environment. 
During the simulation, the driver controls the ego vehicle via a Logitech G27 steering wheel with force feedback, while the longitudinal speed is kept constant at 50km/h. Therefore, the human can control its desired TLC only by the lateral distance and heading error of the vehicle with respect to the road.
The following two main datasets are collected:
\begin{enumerate*}
	\item \textbf{D1}: 10 laps around the track, with a lane width $w = 4.5m$ and asking the driver to just keep the lane, and
	\item \textbf{D2}: 10 laps around the track, with a lane width $w = 8m$ and asking the driver to keep the lane while travelling on the inner radius of the curves.
\end{enumerate*}

\subsection{Vehicle Modelling}\label{vehiclemodellin}
\begin{figure}[!tb]
	\centering
	\def\svgwidth{\columnwidth}
	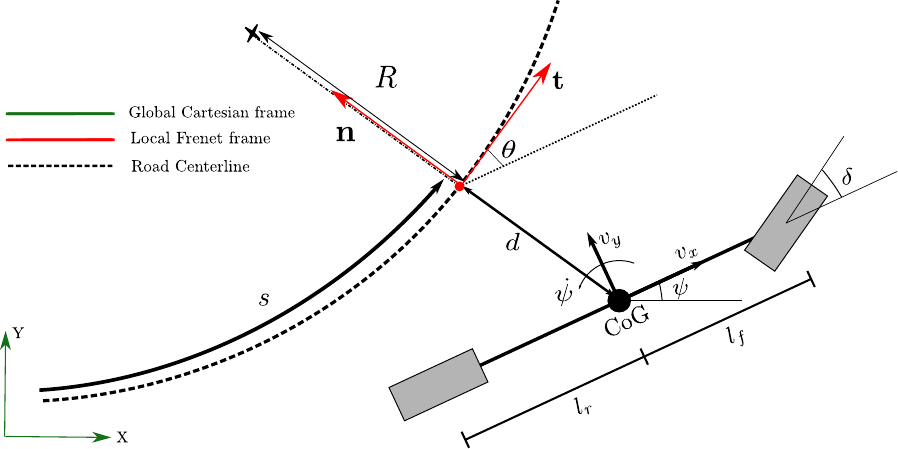
	\caption{Bicycle Model in Frenet coordinates.}
	\label{vehicle}
\end{figure}
The MPC requires a model of the system to be controlled. In our scenario, this model includes the vehicle dynamics and kinematics and the road itself. To do this, we use Frenet coordinates to describe the vehicle kinematic with respect to the centerline of the road. The coordinate system is defined with the following variables, shown in Figure \ref{vehicle}: \begin{enumerate*}
	\item the arc length $s$, representing the travelled distance along the road,
	\item the centerline deviation $d$, representing the lateral signed position on the road with respect to the centerline, according to the $\mathbf{n}-\mathbf{t}$ frame and 
	\item the heading error $\theta$, representing the difference between the yaw angle of the vehicle and the heading of the road.
\end{enumerate*}
The kinematics of the vehicle evolve according to the changing curvature of the road $\kappa(s)$, that we assume as known. Moreover, we assume a constant longitudinal velocity $v_x = 50km/h$. The state equations for the kinematics are \cite{Qian2016}:
\begin{equation}
	\begin{aligned}[t]
		 \dot{s} &= \frac{v_xcos\theta - v_ysin\theta}{1-\kappa(s)d} \\
		 \dot{d} &= v_xsin\theta - v_ycos\theta       \\
		 \dot{\theta} &= \dot{\psi} - \kappa(s)\frac{v_xcos\theta - v_ysin\theta}{1-\kappa(s)d}\;.  \\
	\end{aligned}
\end{equation}
As to the dynamics, we employ a linearized bicycle model, with lateral velocity $v_y$ and yaw rate $\dot{\psi}$ as states \cite{milliken1995race}. The steering angle at the wheel is denoted as $\delta$.

\subsection{MPC Formulation}
Based on the model chosen in the previous section, we can define the states and controls of the MPC as:
\( x = \left( v_y, \dot{\psi}, s, d, \theta, \delta \right)\), \( u = \dot{\delta}\). The control input has been chosen to be the steering rate $\dot{\delta}$ as this is known to improve the smoothness of the control action and allows us to directly impose cost and constraints on its value.
The MPC is formulated as in the following equations:
\begin{align}
		\min_{x,u} \quad & \sum_{k=0}^{N-1} l(\theta, x_k, u_k) + l_N(\theta, x_k, u_k) \label{mpc:cost} \\ 
		\st \quad & x_{k+1} = f(x_k,u_k) \label{mpc:discrmodel} \\
		& -w/2 \leq d \leq w/2  \label{mpc:safetyconstr}\;.
\end{align}
where: eq. (\ref{mpc:discrmodel}) is the Runge-Kutta discretization (sample time of 0.1s) of the vehicle model presented in Section \ref{vehiclemodellin}; eq. (\ref{mpc:safetyconstr}) represents the safety constraints related to the lane boundaries, whose width is $w$; eq. (\ref{mpc:cost}) is the objective function whose parameters are learned with BCO and whose structure is defined for each experiment in Section \ref{results}. The other states and controls are also box-constrained according to physical limits. The control horizon is chosen such that the controller has a lookahead distance of 30m, therefore $N = 20$.
The MPC is implemented using Rockit \cite{gillis2020effortless}, an interface to CasADi \cite{Andersson2018a}, aimed at optimal control problems definition. The problem is then translated into a PyTorch module, by extending its automatic differentiation engine to the one of CasADi. In this way, it is possible to seamlessly use a MPC layer within the PyTorch framework, though there is still the need of moving the tensors onto the CPU. A more efficient implementation of the MPC problem supporting GPU operations may be object of future research.

\subsection{Results}\label{results}
This section shows results of MPC-BCO on the demonstration data D1 and D2 described in Section \ref{demos}. Both datasets are made of 10 laps, denoted as Lap\{i\}. The laps are divided into different trajectories of length T and MPC-BCO is performed with stochastic gradient descent (with Adam algorithm) on batches of 10 trajectories each. The imitation loss $L$ is chosen as the L2 error between the observed and demonstrated lateral centerline deviation, i.e. $d$ and $d^*$ respectively; hence, $L = ||d - d^*||_2^2$. The rollout trajectories are of length $T = 10s$ and $t_s = 5s$ is used for BPTT.

\begin{figure}[!htb]
	\centering
	\resizebox{\columnwidth}{!}{\input{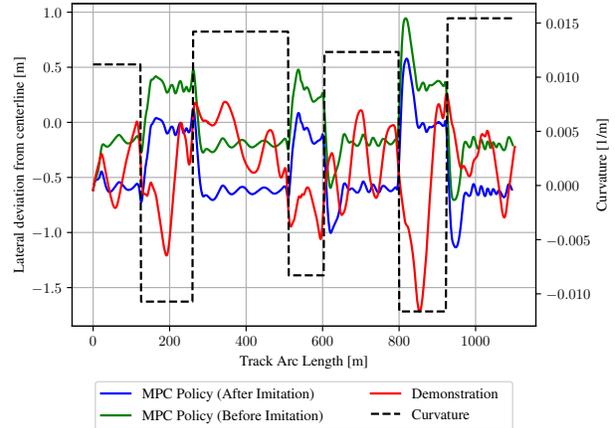}}
	\caption{Learning parameters for $l_{D1}$: demonstrated and learned $d$ on Lap5.}
	\label{staticd}
\end{figure}

Firstly, we learn static parameters in the cost function using the D1 dataset. The designed cost function of the MPC for D1 is $l_{D1} = \sum_{k=0}^{N-1} W_{d}(d_k-\bar{d})^2 + W_{\theta}\theta_k^2 + W_{\dot{\delta}}\dot{\delta}_k^2$, so only a stage cost and no terminal cost involved. The learnable parameters are $(W_{d}, W_{\theta}, W_{\dot{\delta}}, \bar{d})$. To ensure that the cost function is plausible, we project the parameters into a different space. Specifically, the weights $(W_{d}, W_{\theta}, W_{\dot{\delta}})$ are mapped onto a softplus space $\log(1+e^z)$, which guarantees non-negativity, and the offset $\bar{d}$ is mapped onto a $tanh(\bar{d})$ space, multiplied by the limits of the lane boundaries $w/2$. 
The initial parameters are $(1, 1, 1, 0)$. The numerical results in Table \ref{staticparamstable} show a consistent behavior of the learned parameters when learned of different laps; the weight on the lateral centerline deviation is decreased, while the ones on the control amplitude and the heading error are increased. The parameter that influences the most the closed-loop behavior is $\bar{d}$, determining the average preferred lateral centerline offset on the road. Visually, $\bar{d}$ influences the driver perspective on the road during the constant curvature traits 
and therefore also the distance to lane crossing $D$.
In Figure \ref{staticd}, we show a complete lap of the track, perfomed by the human on Lap5 (Demonstration), by the MPC with initial parameters (Before Imitation) and by the MPC with parameters learned on Lap5-6. In the lap performed by the latter, $d$ is shifted towards negative values, i.e. to the right side of the lane, as the demonstrations show this driving preference. 
In Table \ref{baseline_table}, we show some relevant validation metrics compared with a BCO baseline, similarly to \cite{Scheel2021}, where $\pi$ is chosen as a feedforward policy network with two hidden layers of 32 units, as in \cite{Fu2017}. The metrics reflect the main attributes of good human-like autonomous driving policy; first, the imitation performance, computed through $L$, then the safety performance, computed by the number of contraints violations and finally the comfort performance, computed as the integral of the squared lateral acceleration. Their values are computed on and averaged over 5 validation sub-trajectories taken randomly from all the laps. MPC-BCO outperforms BCO on all these attributes; by using the same amount of demonstrations, MPC-BCO can exploit its model and cost to learn with more accuracy than a typical network. Additionaly, thanks to the explicit constraints, MPC-BCO is able to avoid any violation of the lane boundaries, while BCO still violates them quite often. Finally, MPC-BCO improves the comfort of the trajectories; thanks to its cost on the control amplitude, it provides a smoother behavior, with a comfort metric value much closer to the human one, with respect to pure learning-based BCO.
\begin{table}[!htb]
	\centering
	\begin{tabular}{||M{2cm} | M{1cm} M{1cm} M{1cm} M{1cm}||} 
		\hline
		Laps (Sub-datasets) & $W_d$ & $W_{\theta}$ & $W_{\dot{\delta}}$ & $\bar{d}$ \\ 
		\hline\hline
		Lap1-2 & 0.887 & 1.105 & 1.125 & \textbf{-0.191} \\ 
		\hline
		Lap3-4 & 0.839 & 1.181 & 1.181 & \textbf{-0.381}\\
		\hline
		Lap5-6 & 0.875 & 1.127 & 1.121 & \textbf{-0.412}\\
		\hline
		Lap6-8 & 0.829 & 1.209 & 1.196 & \textbf{-0.444} \\
		\hline
	\end{tabular}
	\caption{Learning static parameters for $l_{D1}$: learned values for different subsets of 2 laps each.}
	\label{staticparamstable}
\end{table}
\begin{table}[!htb]
	\centering
	\begin{tabular}{||M{2cm} | M{1.5cm} M{1.5cm}| M{1.5cm} ||} 
		\hline
		\textbf{Metric} & \textbf{BCO} & \textbf{MPC-BCO} & \textbf{Human} \\ 
		\hline\hline
		\emph{Imitation} & 3.0566m & 0.5832m & - \\ 
		\hline
		\emph{Safety} & 90 & 0 & 0\\
		\hline
		\emph{Comfort} & 758.10 & 67.58 & 59.49\\
		\hline
	\end{tabular}
	\caption{Validation results of MPC-BCO compared to BCO. The imitation performance is computed as $L = ||d - d^*||_2^2$, safety by the number of times $|d| > w/2$ and comfort by $\int a_y^2$.}
	\label{baseline_table}
\end{table}

\begin{figure}[!htb]
	\centering
	\resizebox{\columnwidth}{!}{\input{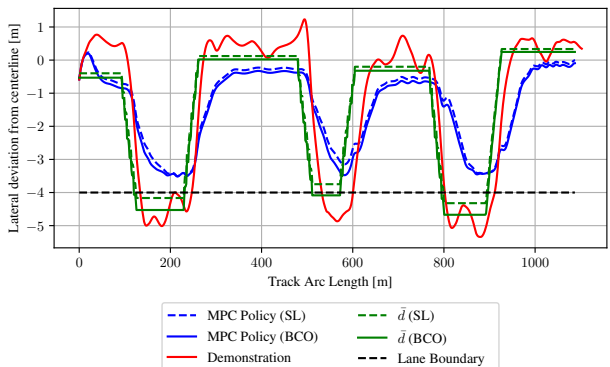}}
	\caption{Learning an indirect controller for $l_{D2}$: demonstrated and learned $d$ (both from SL and MPC-BCO) on Lap5.}
	\label{dynamicd}
\end{figure}

\begin{figure}[!htb]
	\centering
	\resizebox{\columnwidth}{!}{\input{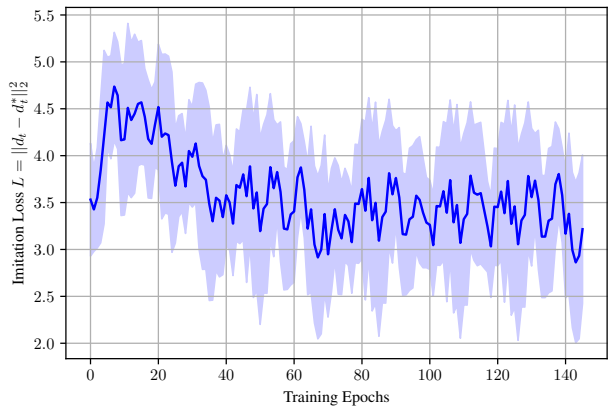}}
	\caption{Convergence result of MPC-BCO. The initial loss is the imitation loss obtained when using $\bar{d}$(SL) into the closed-loop system.}
	\label{loss}
\end{figure}

Secondly, we use demonstrations from D2, which provide a more complex behavior to imitate, depending on the road curvature ahead $\kappa$. Following the findings from the previous experiment, we learn only $\bar{d}$, and set the weights $W_\theta, W_\delta$ equal to 1. Moreover, instead of providing a planned reference for all discretization steps, we consider $\bar{d}$ in the terminal cost as the lateral deviation goal to reach 2s ahead. Hence, we consider the following cost function: $l_{D2} = \sum_{k=0}^{N-1} \theta_k^2 + \dot{\delta}_k^2 + (d_N - \bar{d}(\kappa))^2$.
The indirect controller $\bar{d}(\kappa)$ is chosen to be represented as a piecewise linear function, approximated as a multilayer perceptron (MLP) with two layers of 50 hidden neurons, with a rectifying linear unit (ReLu) in between. Its inputs are chosen to be seven road curvature values, starting from the current one up to 30m ahead, equally spaced of 5m between each other. 
As starting point, $\bar{d}(\kappa)$ is learned in open-loop. The laps are divided into trajectories of 2s each and supervised learning (SL) is done, according to a mean squared error, to match $\bar{d}(\kappa)$ to the final $d$ in the 2s trajectory, given the $\kappa$ inferred from the initial state of the trajectory. Then, as in the previous experiments, laps are again divided into 10s trajectories and MPC-BCO is run to refine the MLP parameters optimizing the closed-loop performance. In Figure \ref{loss}, we show the convergence results of this learning, starting from the imitation loss that is obtained when using $\bar{d}(\kappa)$ learned with SL in the closed-loop system. From that point, MPC-BCO reduces the imitation loss by 50\%, by taking into consideration the closed-loop effects.
In Figure \ref{dynamicd} we show a complete lap of the track, perfomed by the human on Lap5 (Demonstration), by the MPC with $\bar{d}(\kappa)$ learned with SL and by the MPC with $\bar{d}(\kappa)$ learned with MPC-BCO.
It is interesting to point out that the more the learned reference violates the safety boundary, the more the imitation loss on the closed-loop trajectories improve, though staying safe. This hints that, for a better closed-loop balance between safety and performance, the indirect controller should not be constrained on safety. This also highlights the difference to other approaches that learn a safe planning module followed by an idealized tracking controller, whereas MPC-IL leverages on the end-to-end differentiability of the policy to directly optimize the closed-loop behavior through MPC parameters.

\section{Conclusions and Future Works}
In this paper we have explored the use of MPC as a controller within an autonomous driving policy, learned via imitation learning. Firstly, we have outlined the methodology of MPC-IL consisting of a policy with a MPC reactive controller, preceded by indirect imitation learning-based controllers that based on the current latent information, output the optimal parameters in the MPC cost function, model or constraints. A novelty of this approach lies in the combination of both learning-based indirect controllers and the end-to-end differentiability of the framework, possible by leveraging on implicit differentiation. In this way, the safety of the MPC control and the flexibility of end-to-end learning can be combined in closed-loop imitation learning algorithms for autonomous driving. Preliminary experimental results have been shown for a lane keeping controller.
This is an introductory work and it has its limitations; for example, the experimental results were obtained considering a simple, non-interactive and artificially constructed scenario where a predefined structure of the MPC cost function and the behavioral cloning loss were possible. Moreover, we see potential in using camera images as input to the indirect controllers, in order to compare its performance and safety with respect to popular end-to-end controllers for AVs, directly mapping images to control actions.

\section*{Acknowledgements}
This work was supported by the Flemish Agency for Innovation and Entrepreneurship (VLAIO) in the context of MIMIC (huMan IMItation for autonomous driving Comfort) Baekeland Mandaat [Project nr. HBC.2020.2263].
\bibliography{example_paper}

\begin{thebibliography}{30}
\providecommand{\natexlab}[1]{#1}
\providecommand{\url}[1]{\texttt{#1}}
\expandafter\ifx\csname urlstyle\endcsname\relax
  \providecommand{\doi}[1]{doi: #1}\else
  \providecommand{\doi}{doi: \begingroup \urlstyle{rm}\Url}\fi

\bibitem[Acerbo et~al.(2020)Acerbo, der Auweraer, and Son]{Acerbo2020}
Acerbo, F.~S., der Auweraer, H.~V., and Son, T.~D.
\newblock Safe and computational efficient imitation learning for autonomous
  vehicle driving.
\newblock In \emph{2020 American Control Conference ({ACC})}. {IEEE}, jul 2020.
\newblock \doi{10.23919/acc45564.2020.9147256}.

\bibitem[Amos et~al.(2018)Amos, Jimenez, Sacks, Boots, and Kolter]{Amos2018}
Amos, B., Jimenez, I., Sacks, J., Boots, B., and Kolter, J.~Z.
\newblock Differentiable mpc for end-to-end planning and control.
\newblock \emph{Advances in neural information processing systems}, 31, 2018.

\bibitem[Andersson \& Rawlings(2018)Andersson and Rawlings]{Andersson2018}
Andersson, J.~A. and Rawlings, J.~B.
\newblock Sensitivity analysis for nonlinear programming in {CasADi}.
\newblock \emph{{IFAC}-{PapersOnLine}}, 51\penalty0 (20):\penalty0 331--336,
  2018.
\newblock \doi{10.1016/j.ifacol.2018.11.055}.

\bibitem[Andersson et~al.(2018)Andersson, Gillis, Horn, Rawlings, and
  Diehl]{Andersson2018a}
Andersson, J. A.~E., Gillis, J., Horn, G., Rawlings, J.~B., and Diehl, M.
\newblock {CasADi}: a software framework for nonlinear optimization and optimal
  control.
\newblock \emph{Mathematical Programming Computation}, 11\penalty0
  (1):\penalty0 1--36, jul 2018.
\newblock \doi{10.1007/s12532-018-0139-4}.

\bibitem[Bansal et~al.(2018)Bansal, Krizhevsky, and Ogale]{Bansal2018}
Bansal, M., Krizhevsky, A., and Ogale, A.
\newblock Chauffeurnet: Learning to drive by imitating the best and
  synthesizing the worst.
\newblock \emph{arXiv preprint arXiv:1812.03079}, 2018.

\bibitem[Barendswaard et~al.(2019)Barendswaard, van Breugel, Schelfaut,
  Sluijter, Zuiker, Pool, Boer, and Abbink]{Barendswaard2019}
Barendswaard, S., van Breugel, L., Schelfaut, B., Sluijter, J., Zuiker, L.,
  Pool, D.~M., Boer, E.~R., and Abbink, D.~A.
\newblock Effect of velocity and curve radius on driver steering behaviour
  before curve entry.
\newblock In \emph{2019 {IEEE} International Conference on Systems, Man and
  Cybernetics ({SMC})}. {IEEE}, oct 2019.
\newblock \doi{10.1109/smc.2019.8914263}.

\bibitem[Bellem et~al.(2016)Bellem, Schönenberg, Krems, and
  Schrauf]{Bellem2016}
Bellem, H., Schönenberg, T., Krems, J.~F., and Schrauf, M.
\newblock Objective metrics of comfort: Developing a driving style for highly
  automated vehicles.
\newblock \emph{Transportation Research Part F: Traffic Psychology and
  Behaviour}, 41:\penalty0 45--54, aug 2016.
\newblock \doi{10.1016/j.trf.2016.05.005}.

\bibitem[Blockwitz et~al.(2012)Blockwitz, Otter, Akesson, Arnold, Clauss,
  Elmqvist, Friedrich, Junghanns, Mauss, Neumerkel, Olsson, and
  Viel]{Blockwitz2012}
Blockwitz, T., Otter, M., Akesson, J., Arnold, M., Clauss, C., Elmqvist, H.,
  Friedrich, M., Junghanns, A., Mauss, J., Neumerkel, D., Olsson, H., and Viel,
  A.
\newblock Functional mockup interface 2.0: The standard for tool independent
  exchange of simulation models.
\newblock In \emph{Linköping Electronic Conference Proceedings}. Linköping
  University Electronic Press, nov 2012.
\newblock \doi{10.3384/ecp12076173}.

\bibitem[Chen et~al.(2019)Chen, Yuan, and Tomizuka]{Chen2019}
Chen, J., Yuan, B., and Tomizuka, M.
\newblock Deep imitation learning for autonomous driving in generic urban
  scenarios with enhanced safety.
\newblock In \emph{2019 {IEEE}/{RSJ} International Conference on Intelligent
  Robots and Systems ({IROS})}. {IEEE}, nov 2019.
\newblock \doi{10.1109/iros40897.2019.8968225}.

\bibitem[Elbanhawi et~al.(2015)Elbanhawi, Simic, and Jazar]{Elbanhawi2015}
Elbanhawi, M., Simic, M., and Jazar, R.
\newblock In the passenger seat: Investigating ride comfort measures in
  autonomous cars.
\newblock \emph{{IEEE} Intelligent Transportation Systems Magazine}, 7\penalty0
  (3):\penalty0 4--17, 2015.
\newblock \doi{10.1109/mits.2015.2405571}.

\bibitem[Fu et~al.(2017)Fu, Luo, and Levine]{Fu2017}
Fu, J., Luo, K., and Levine, S.
\newblock Learning robust rewards with adversarial inverse reinforcement
  learning.
\newblock \emph{arXiv preprint arXiv:1710.11248}, 2017.

\bibitem[Ghasemipour et~al.(2020)Ghasemipour, Zemel, and Gu]{Ghasemipour2019}
Ghasemipour, S. K.~S., Zemel, R., and Gu, S.
\newblock A divergence minimization perspective on imitation learning methods.
\newblock In Kaelbling, L.~P., Kragic, D., and Sugiura, K. (eds.),
  \emph{Proceedings of the Conference on Robot Learning}, volume 100 of
  \emph{Proceedings of Machine Learning Research}, pp.\  1259--1277. PMLR, 30
  Oct--01 Nov 2020.

\bibitem[Gillis et~al.(2020)Gillis, Vandewal, Pipeleers, and
  Swevers]{gillis2020effortless}
Gillis, J., Vandewal, B., Pipeleers, G., and Swevers, J.
\newblock Effortless modeling of optimal control problems with rockit.
\newblock In \emph{39th Benelux Meeting on Systems and Control, Date:
  2020/03/10-2020/03/12, Location: Elspeet, The Netherlands}, 2020.

\bibitem[Godthelp et~al.(1984)Godthelp, Milgram, and Blaauw]{Godthelp1984}
Godthelp, H., Milgram, P., and Blaauw, G.~J.
\newblock The development of a time-related measure to describe driving
  strategy.
\newblock \emph{Human Factors: The Journal of the Human Factors and Ergonomics
  Society}, 26\penalty0 (3):\penalty0 257--268, jun 1984.
\newblock \doi{10.1177/001872088402600302}.

\bibitem[Gros \& Zanon(2020)Gros and Zanon]{Gros2020}
Gros, S. and Zanon, M.
\newblock Data-driven economic {NMPC} using reinforcement learning.
\newblock \emph{{IEEE} Transactions on Automatic Control}, 65\penalty0
  (2):\penalty0 636--648, feb 2020.
\newblock \doi{10.1109/tac.2019.2913768}.

\bibitem[Hatz et~al.(2012)Hatz, Schlöder, and Bock]{Hatz2012}
Hatz, K., Schlöder, J.~P., and Bock, H.~G.
\newblock Estimating parameters in optimal control problems.
\newblock \emph{{SIAM} Journal on Scientific Computing}, 34\penalty0
  (3):\penalty0 A1707--A1728, jan 2012.
\newblock \doi{10.1137/110823390}.

\bibitem[Hawke et~al.(2020)Hawke, Shen, Gurau, Sharma, Reda, Nikolov, Mazur,
  Micklethwaite, Griffiths, Shah, and Kndall]{Hawke2020}
Hawke, J., Shen, R., Gurau, C., Sharma, S., Reda, D., Nikolov, N., Mazur, P.,
  Micklethwaite, S., Griffiths, N., Shah, A., and Kndall, A.
\newblock Urban driving with conditional imitation learning.
\newblock In \emph{2020 {IEEE} International Conference on Robotics and
  Automation ({ICRA})}. {IEEE}, may 2020.
\newblock \doi{10.1109/icra40945.2020.9197408}.

\bibitem[Heess et~al.(2015)Heess, Wayne, Silver, Lillicrap, Erez, and
  Tassa]{Heess2015}
Heess, N., Wayne, G., Silver, D., Lillicrap, T., Erez, T., and Tassa, Y.
\newblock Learning continuous control policies by stochastic value gradients.
\newblock \emph{Advances in neural information processing systems}, 28, 2015.

\bibitem[Kalabic et~al.(2019)Kalabic, Chakrabarty, Quirynen, and
  Cairano]{Kalabic2019}
Kalabic, U., Chakrabarty, A., Quirynen, R., and Cairano, S.~D.
\newblock Learning autonomous vehicle passengers' preferred driving styles
  using g-g plots and haptic feedback.
\newblock In \emph{2019 {IEEE} Intelligent Transportation Systems Conference
  ({ITSC})}. {IEEE}, oct 2019.
\newblock \doi{10.1109/itsc.2019.8917328}.

\bibitem[Kolekar et~al.(2020)Kolekar, de~Winter, and Abbink]{Kolekar2020}
Kolekar, S., de~Winter, J., and Abbink, D.
\newblock Human-like driving behaviour emerges from a risk-based driver model.
\newblock \emph{Nature Communications}, 11\penalty0 (1), sep 2020.
\newblock \doi{10.1038/s41467-020-18353-4}.

\bibitem[Lazcano et~al.(2021)Lazcano, Niu, Akutain, Cole, and
  Shyrokau]{Lazcano2021}
Lazcano, A.~M., Niu, T., Akutain, X.~C., Cole, D., and Shyrokau, B.
\newblock {MPC}-based haptic shared steering system: A driver modeling approach
  for symbiotic driving.
\newblock \emph{{IEEE}/{ASME} Transactions on Mechatronics}, 26\penalty0
  (3):\penalty0 1201--1211, jun 2021.
\newblock \doi{10.1109/tmech.2021.3063902}.

\bibitem[Martinsen et~al.(2020)Martinsen, Lekkas, and Gros]{Martinsen2020}
Martinsen, A.~B., Lekkas, A.~M., and Gros, S.
\newblock Combining system identification with reinforcement learning-based
  mpc.
\newblock \emph{IFAC-PapersOnLine}, 53\penalty0 (2):\penalty0 8130--8135, 2020.

\bibitem[Milliken et~al.(1995)Milliken, Milliken, et~al.]{milliken1995race}
Milliken, W.~F., Milliken, D.~L., et~al.
\newblock \emph{Race car vehicle dynamics}, volume 400.
\newblock Society of Automotive Engineers Warrendale, PA, 1995.

\bibitem[Nash \& Cole(2019)Nash and Cole]{Nash2019}
Nash, C.~J. and Cole, D.~J.
\newblock Identification and validation of a driver steering control model
  incorporating human sensory dynamics.
\newblock \emph{Vehicle System Dynamics}, 58\penalty0 (4):\penalty0 495--517,
  mar 2019.
\newblock \doi{10.1080/00423114.2019.1589536}.

\bibitem[Pan et~al.(2017)Pan, Cheng, Saigol, Lee, Yan, Theodorou, and
  Boots]{Pan2017}
Pan, Y., Cheng, C.-A., Saigol, K., Lee, K., Yan, X., Theodorou, E., and Boots,
  B.
\newblock Agile autonomous driving using end-to-end deep imitation learning.
\newblock \emph{arXiv preprint arXiv:1709.07174}, 2017.

\bibitem[Pulver et~al.(2021)Pulver, Eiras, Carozza, Hawasly, Albrecht, and
  Ramamoorthy]{Pulver2021}
Pulver, H., Eiras, F., Carozza, L., Hawasly, M., Albrecht, S.~V., and
  Ramamoorthy, S.
\newblock {PILOT}: Efficient planning by imitation learning and optimisation
  for safe autonomous driving.
\newblock In \emph{2021 {IEEE}/{RSJ} International Conference on Intelligent
  Robots and Systems ({IROS})}. {IEEE}, sep 2021.
\newblock \doi{10.1109/iros51168.2021.9636862}.

\bibitem[Qian et~al.(2016)Qian, de~La~Fortelle, and Moutarde]{Qian2016}
Qian, X., de~La~Fortelle, A., and Moutarde, F.
\newblock A hierarchical model predictive control framework for on-road
  formation control of autonomous vehicles.
\newblock In \emph{2016 {IEEE} Intelligent Vehicles Symposium ({IV})}. {IEEE},
  jun 2016.
\newblock \doi{10.1109/ivs.2016.7535413}.

\bibitem[Scheel et~al.(2021)Scheel, Bergamini, Wolczyk, Osinski, and
  Ondruska]{Scheel2021}
Scheel, O., Bergamini, L., Wolczyk, M., Osinski, B., and Ondruska, P.
\newblock Urban driver: Learning to drive from real-world demonstrations using
  policy gradients.
\newblock In \emph{Conference on Robot Learning (CoRL)}, 2021.

\bibitem[Suo et~al.(2021)Suo, Regalado, Casas, and Urtasun]{Suo2021}
Suo, S., Regalado, S., Casas, S., and Urtasun, R.
\newblock Trafficsim: Learning to simulate realistic multi-agent behaviors.
\newblock In \emph{Proceedings of the IEEE/CVF Conference on Computer Vision
  and Pattern Recognition}, pp.\  10400--10409, 2021.

\bibitem[Torabi et~al.(2018)Torabi, Warnell, and Stone]{Torabi2018}
Torabi, F., Warnell, G., and Stone, P.
\newblock Behavioral cloning from observation.
\newblock \emph{arXiv preprint arXiv:1805.01954}, 2018.

\end{thebibliography}


\begin{thebibliography}{0}
\providecommand{\natexlab}[1]{#1}
\providecommand{\url}[1]{\texttt{#1}}
\expandafter\ifx\csname urlstyle\endcsname\relax
  \providecommand{\doi}[1]{doi: #1}\else
  \providecommand{\doi}{doi: \begingroup \urlstyle{rm}\Url}\fi

\end{thebibliography}
\bibliographystyle{icml2022}


\end{document}